\documentclass[copyright,noncommercial]{eptcs}
\pdfoutput=1
\providecommand{\volume}{5}
\providecommand{\anno}{2009}
\providecommand{\firstpage}{5}
\providecommand{\eid}{1}
\providecommand{\event}{Y. Deville and C. Solnon (Eds): Sixth International Workshop\\
 on Local Search Techniques in Constraint Satisfaction (LSCS'09).}

\usepackage{graphics}
 \usepackage{graphicx}
 \usepackage{multirow}
 \usepackage{color}
 \usepackage[ruled,vlined,linesnumbered]{algorithm2e}
 \usepackage{algorithmic}
 \usepackage{multirow}

 \usepackage{amsmath}
 \usepackage{amssymb}
 \usepackage{amsmath, amsthm, amssymb}

 \title{A Local Search Modeling for Constrained Optimum Paths Problems (Extended Abstract)}
 \author{Quang Dung Pham
 \institute{
          Universit\'{e} catholique de Louvain\\
         B-1348 Louvain-la-Neuve, Belgium}
 \email{quang.pham@uclouvain.be}
 \and
 Yves Deville
 \institute{
          Universit\'{e} catholique de Louvain\\
         B-1348 Louvain-la-Neuve, Belgium}
 \email{yves.deville@uclouvain.be}
 \and
 Pascal van Hentenryck
 \institute{Brown University, Box 1910\\
         Providence, RI 02912, USA}
 \email{pvh@cs.brown.edu}
 }

 \def\titlerunning{A Local Search Modeling for Constrained Optimum Paths Problems}
 \def\authorrunning{Q. D. Pham, Y. Deville, and P. van Hentenryck}
 \begin{document}
 \maketitle

 \begin{abstract}
 Constrained Optimum Path (COP) problems appear in many real-life applications, especially on communication networks. Some of these problems have been considered and solved by specific techniques which are usually difficult to extend. In this paper, we introduce a novel local search modeling for solving some COPs by local search. The modeling features the compositionality, modularity, reuse and strengthens the benefits of Constrained-Based Local Search \cite{pvh05}. We also apply the modeling to the edge-disjoint paths problem (EDP). Computational results show the significance of the approach.
 \end{abstract}

 \section{Introduction}

 Constrained Optimum Path (COP) problems, where optimum path from origin to destination satisfying additional constraints must be found, appear in many real-life applications, especially on communication and transportation networks. These problems have attracted considerable attention from different research communities: operations research, telecommunications because of its various applications (see \cite{Carlyle03} and the references therein).

 Most of COP problems are NP-hard. Some instances can be solved efficiently by specific techniques, for instance, branch and bound using a Lagrangian-based bound \cite{beasley89}, vertex-labeling \cite{dumitrescu03}, etc. These techniques seem to be sophisticated and depend on particular constraints and objective functions to be optimized. Moreover, they are difficult to extend, for example, when we face with generalized problems where more constraints are required to be satisfied. We propose in this paper a novel approach for modeling and solving some COP problems by local search where the desired paths are elementary (i.e. no repeated nodes). The objective of this work is to extend the \texttt{LS(Graph \& Tree)} framework \cite{pqdung09} by the design and the implementation of abstractions allowing to easily model and solve some COP problems using this approach. The computational model features compositionality, modularity, reuse, and strengthens the local search modeling benefits of Constraint-Based Local Search \cite{pvh05} which provides the separation of concerns.

 The proposed computational model has been applied to the EDP problem. Computational results show the significance of the approach.

 \section{Modeling paths with spanning trees}
 We introduce in this section the proposed approach for modeling COP problems with spanning trees. We first give somes definitions and notations over graphs. Our framework considers both directed and undirected graphs but for simplicity, we consider in this presentation only undirected graphs.

 Given  an undirected graph $g$, we denote $V(g)$, $E(g)$ respectively the set of nodes and the set of edges of $g$.

 In this paper, we only consider elementary paths, henceforth we use the word "path" instead of "elementary path" if there is no ambiguity. A graph is connected if and only if there exists a path from $u$ to $v, \forall u,v\in V(g)$. A tree is an undirected connected graph containing no cycles. A spanning tree $tr$ of an undirected connected graph $g$ is a tree spanning all the nodes of $g$: $V(tr)=V(g)$ and $E(tr)\subseteq E(g)$. A tree \textit{tr} is called a rooted tree at $r$ if the node $r$ has been designated the root. Each edge of $tr$ is implicitly oriented towards the root. If the edge $(u,v)$ is oriented from $u$ to $v$, we call $v$ the father of $u$ on $tr$.

 The key decision design is inspired from the following observation: Given a rooted tree $tr$ whose root is $t$, the path from a given node $s$ to $t$ on $tr$ is unique. An update of $tr$ will generate a new rooted tree which may induce a new path from $s$ to $t$ on this tree.

 Given an undirected graph $g$ and a node $r\in V(g)$, \textit{VarRootedSpanningTree}$(g,r)$ (also called rooted spanning tree variable) is a concept representing a dynamic spanning tree of $g$. The spanning tree is rooted at $r$.

 In order to model a COP problem in an undirected graph $g$ in which the source and the target of the desired path are respectively $s,t\in V(g)$, we use \textit{VarRootedSpanningTree}$(g,s,t)$ which is \textit{VarRootedSpanningTree}$(g,t)$ with a node $s$ designated as the source node. Each instance $tr$ of \textit{VarRootedSpanningTree}$(g,s,t)$ specifies a unique path from $s$ to $t$ on $g$. Henceforth we use $s$ to denote the source node of any rooted spanning tree of the given graph and the path from $s$ to the root of a rooted spanning tree $tr$ on $tr$ is called the path induced by $tr$ if there is no ambiguity.

 The main avantage of using rooted spanning tree for modeling paths instead of using explicit paths representation (i.e. a sequence of nodes) is the simplification of neighborhood computation. The tree structure constains rich information that induces directly path structure from a node $s$ to the root. A simple update over that tree (i.e. an edge replacement which is detailed in Section \ref{neighborhood}) will induce a new path from $s$ to the root.

 \section{Neighborhood}\label{neighborhood}
 Given an instance $tr$ of \textit{VarRootedSpanningTree}$(g,s,t)$, we show how to change $tr$ in order to generate a new rooted spanning tree $tr'$ of $g$ which induces a new path from $s$ to $t$ on $g$.

 Given an undirected graph $g$, an instance $tr$ of \textit{VarRootedSpanningTree}$(g,s,t)$, an edge $e=(u,v)$ such that $e\in E(g)\setminus E(tr)$ is called \textit{replacing} edge of $tr$. We denote $rpl(tr)$ the set of \textit{replacing} edges of $tr$. Given $e\in rpl(tr)$, an edge $e'$ that belongs to the path between two endpoints of $e$ on $tr$ is called \textit{replacable} edge of $e$. We denote $rpl(tr,e)$ the set of \textit{replacable} edges of $e$. Intuitionally, a \textit{replacing} edge $e$ is an edge that is not in the tree $tr$ but that can be added to $tr$ (this edge insertion creates a cycle $C$ when we ignore orientations of edges of $tr$), and all edges of this cycle except $e$ are \textit{replacable} edges of $e$.

 Given an undirected graph $g$, an instance $tr$ of \textit{VarRootedSpanningTree}$(g,s,t)$, $e$ and $e'$ are respectively \textit{replacing} edge of $tr$ and \textit{replacable} edge of $e$, we define the following edge replacement action:
 \begin{enumerate}
     \item Insert the edge $e=(u,v)$ to $tr$. This creates an undirected graph $g$ with a cycle $C$ containing the edge $e'$.
     \item Remove $e'$ from $g$.
 \end{enumerate}

 After taking above edge replacement action, we obtain a new rooted spaning tree $tr'$ of $g$. We denote $tr'=rep(tr,e',e)$. The neighborhood of $tr$ is \[N(tr)=\{tr'=rep(tr,e',e)\mid e\in rpl(tr),e'\in rpl(tr,e)\}\]

 It is easy to observe that two different spanning trees $tr_1$ and $tr_2$ rooted at $t$ of an undirected graph $g$ may induce the same path from $s$ to $t$ on $g$ ($s,t\in V(g)$). The neighborhood $N(tr)$ must then be reduced such that the new tree induces a new path from $s$ to $t$. This reduction is described in \cite{techrep09}. The action $rep(tr,eo,ei)$ is called a basic move. Figure 1 gives an example of basic move.

 \begin{figure}\label{basicmove}
 \begin{center}
 \begin{tabular}{cc}
 \includegraphics[width=3cm]{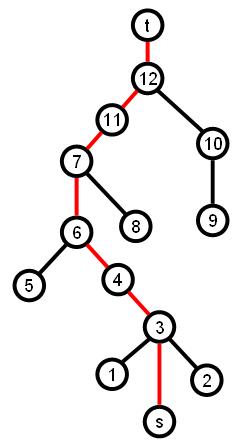} & \includegraphics[width=3cm]{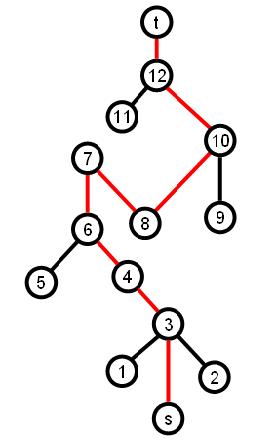}\\
 a. current tree $tr$ & b. $tr'=rep(tr,(7,11),(8,10))$
 \end{tabular}
 \end{center}
 \caption{Example of basic move}
 \end{figure}

 It is possible to consider more complex moves by applying a set of independent basic moves. Two basic moves are independent if the execution of the first one does not affect the second one and vice versa. The execution order of these basic moves does not affect the final result. Figure 2 gives an example of complex move.

 \begin{figure}\label{complexmove}
 \begin{center}
 \begin{tabular}{cc}
 \includegraphics[width=3cm]{originaltree.jpg} & \includegraphics[width=3cm]{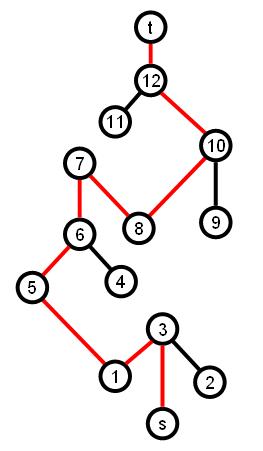}\\
 a. current tree $tr$ & b. $tr'=rep(tr,(7,11),(8,10),(3,4),(1,5))$
 \end{tabular}
 \end{center}
 \caption{Example of complex move}
 \end{figure}

 \section{\texttt{COMET} implementation}
 We extend the \texttt{LS(Graph \& Tree)} framework by implementing some \textit{GraphInvariants}, \textit{GraphConstraint}s and \textit{GraphObjective}s (see \cite{pqdung09} for more detail) for modeling and solving some COP problems. \textit{GraphInvariant} is a concept representing objects which maintain some properties of a dynamic graph\footnote{dynamic graph is a graph that can be changed e.g., by the removal or the insertion of vertices, edges.} (for instance, the sum of weights of all the edges of a graph, the diameter of a tree, etc.). \textit{GraphConstraint} and \textit{GraphObjective} are concepts describing differentiable objects which maintain some properties (for instance, the number of violations of a constraint or the value of an objective function) of a dynamic graph. The main feature of \textit{GraphConstraint} and \textit{GraphObjective} is the interface, allowing to query the impact of local moves (modification of the dynamic graph) on these properties. Some \textit{GraphConstraints} and \textit{GraphObjectives} have been designed and implemented over \textit{VarRootedSpanningTree}. For instance, \texttt{PathCostOnEdges(tr,k)} \footnote{\texttt{tr} is a \textit{VarRootedSpanningTree}, \texttt{k} is the index of the considered weight on edges.} is an abstraction representing the total weight accumulated along the path induced by \texttt{tr}, \texttt{MinEdge\-Cost(tr,k)}, \texttt{MaxEdgeCost(tr,k)} are abstractions representing the minimal and maximal weight of edges on the path induced by \texttt{tr}. \texttt{NodesVisited(tr,S)} is an abstraction representing the number of nodes of \texttt{S} visited by the path induced by \texttt{tr}. These abstractions are example of \textit{GraphObjectives} which are fundamental when modeling Constrained Optimum Path problems. For example, in QoS, we consider shortest path from an origin to a destination with constraints over bandwidth which is defined to be the minimum weight of edges on the specified path. The framework provides flexibility for modeling various Constrained Optimum Path problems. We can easily combine (with +,-,* operator) and state basic constraints (for instance, $<=, >=, ==$) over these abstractions. \texttt{PathEdgeDisjoint(\-tr)} is another \textit{GraphConstraint} which is defined over an array of paths (the $i^{th}$ path is induced by \texttt{tr[i]}) and specifies that these paths are mutually edge-disjoint.

In order to illustrate the modeling, we give a snippet (see Figure 3) which explores the basic neighborhood. Line 1 initializes a \texttt{LSGraphSolver} object \texttt{ls} which manages all the
\textit{VarGraph}, \textit{VarTree}, \textit{GraphInviant}s,
\textit{GraphConstraint}s and \textit{GraphObjective}s and relations
(dependency graph) between these objects. Line 2 declares and initializes randomly a \texttt{VarRootedSpanningTree} \texttt{tr} rooted at \texttt{t} of the input graph \texttt{g} which represents the path from the source node \texttt{s} to \texttt{t}. \texttt{prefReplacing} and \texttt{prefReplacable} are \textit{GraphInvariants} which maintain the set of \textit{preferred replacing} edges\footnote{edges used for the edge replacements which return new trees inducing new paths from $s$ to $t$.} and \textit{preferred replacable} edges (lines 3-4). Lines 5-7 explore the basic neighborhood and evaluate the quality of moves with respect to a \textit{GraphConstraint} \texttt{C}. The \texttt{getReplaceEdgeDelta} (line 7) method returns the variation of the number of violations of \texttt{C} when the \textit{preferred replacable} edge \texttt{eo} is replaced by the \textit{preferred replacing} edge \texttt{ei} on \texttt{tr}.

\begin{figure}\label{modelingexample}
\texttt{
\begin{tabbing}
1. LSGraphSolver ls();\\
2. VarR\=ootedSpanningTree tr(ls,g,s,t);\\
3. PreferredReplacingEdges prefReplacing(tr);\\
4. PreferredReplacableEdges prefReplacable(tr);\\
5. fora\=ll(ei in prefReplacing.getSet())\\
6. \>fora\=ll(eo in prefReplacable.getSet(ei))\\
7. \>\>d = C.getReplaceEdgeDelta(tr,eo,ei);
\end{tabbing}
}
\caption{Exploring the basic neighborhood}
\end{figure}

\section{Application: The EDP problem}

Given an undirected graph $G=(V,E)$, and a set $T=\lbrace<s_i,t_i>\mid s_i\neq t_i\in V\rbrace$ representing a list of commodities ($\sharp T=k$). EDP consists of finding a maximal cardinality set of mutually edge-disjoint paths from $s_i$ to $t_i$ on $G$ ($<s_i,t_i>\in T$). In \cite{blesa07}, a Multi-start Simple Greedy algorithm is presented as well as an ACO algorithm. The ACO is known to be state-of-the-art for this problem. We propose a local search algorithm using the modeling approach. The model is given in Figure 4 where line 2 initializes an array of \texttt{k} \texttt{VarRootedSpanningTree}s representing \texttt{k} paths between commodities. The edge-disjoint constraint \texttt{ed} is which is defined over paths from \texttt{s[i]} to \texttt{t[i]} on \texttt{tr[i]} (\texttt{i = 1, 2, ..., k}) stated in line 3.
\begin{figure}
\texttt{
\begin{tabbing}
void \=stateModel\{\\
1. \>LSGraphSolver ls();\\
2. \>VarRootedSpanningTree tr[i in 1..k](ls,g,s[i],t[i]);\\
3. \>PathEdgeDisjoint ed(tr);\\
4.\>ls.close();\\
5.\}
\end{tabbing}
}
\label{edp_model}
\caption{The Modeling for the EDP problem}
\end{figure}
In \cite{blesa07}, the following criterion is introduced which quantifies the degree of non-disjointness of a solution $S=\{P_1,P_2,...P_k\}$ ($P_j$ is a path from $s_j$ to $t_j$):
\[C(S)=\sum_{e\in E}(max\{0,\sum_{P_j\in S}\rho^j(S,e)-1\})\] where $\rho^j(S,e)=1$, if $e\in P_j\in S$ and $\rho^j(S,e)=0$, otherwise.

The number of violations of the $PathEdgeDisjoint(P_1,P_2,...,P_k)$ constraint in the framework is defined to be $C(\{P_1,P_2,...,P_k\})$ and the proposed local search algorithm tries to minimize this criterion. From a solution which is normally a set of $k$ non-disjoint, a feasible solution to the EDP problem can be extracted by iteratively removing the path which has most edges in common with other paths until all remaining paths are mutually edge-disjoint as suggested in \cite{blesa07}. In our local search model, we extend that idea by taking a simple greedy algorithm over the remaining paths after that extraction procedure in hope of improving the number of edge-disjoint paths.

The main idea for the search is to try different moves to get a first improvement: 1-move or 2-move over one \texttt{VarRootedSpanningTree}, two 1-moves at hand over two \texttt{VarRootedSpanningTree}s.

For the experimentation, we re-implemented in \texttt{COMET} the Multi-start Greedy Algorithm (MSGA) and the ACO (the extended version) algorithm which are described in \cite{blesa07}, and compare them with our local search model. The instances experimented (graphs including commodities) in the original paper \cite{blesa07} is not available at the moment (except some graphs). We base on the instances generation description in the paper \cite{blesa07} and generate new instances as follows. We take 4 graphs from \cite{blesa07}. For each graph, we generate randomly different sets of commodities with different sizes depending on the size of the graph: for each graph of size $n$, we generate randomly 20 instances with 0.10*$n$, 0.25*$n$ and 0.40*$n$ commodities. In total, we have 240 problems instances. Due to the huge complexity of the problem, we execute each problem instance once with a time limit of 30 minutes for each execution. Experimental results are shown in Table 1. The time window for the MSGA and the ACO algorithms are also 30 minutes. The Table reports the average values of the objective function of the best solutions found and the average of times for obtaining these best solutions of 20 instances (a graph $G=(V,E)$ and a set of $r*|V|$ commodities, $r=0.10,0.25,0.40$). The Table shows that our local search model gives competitive results in comparison with the MSGA and the ACO algorithms. In comparison with the MSGA, our local search model find better solutions in 217/240 instances while the MSGA find better solutions in 4/240 instances. On the other hand, in comparison with the ACO model, our local search model find better solutions in 144/240 instances while the ACO model find better solutions in 11/240 instances.

\section{Conclusion and future work}
We introduce in this paper a novel local search modeling for Constrained Optimum Path problems on graphs. The objective here is to give a high-level modeling framework for implementing some COP problems which strengthens the benefits modeling of CBLS and features compositionality, modularity and reuse. The modeling provides a clean seperation of concerns: The modeling component and the search component are independent. On one hand, it is easy to add new constraints and to modify or remove existing ones, without having to worry about the global effect of these changes. On the other hand, programmers can experiment with different heuristic and metaheuristics without affecting the problem modeling. The modeling is based on $VarRootedSpanningTree(g,s,t)$ concept inspiring the observation that each tree induces a unique path between two specified nodes and a update over this tree satisfying additional contraints generates a new tree which induces a new path between these nodes. Various neighborhoods for have been defined. The abstraction is implemented by extending the \texttt{LS(Graph \& Tree)} in \texttt{COMET}. The modeling has been experimented on the resources constrained shortest path problem and the edge-disjoint paths problem which show the significance of the framework.


\begin{table*}[t]\label{edp}
\begin{small}{
\begin{center}
\begin{tabular}{|l|l|r|r|r|r|r|r|}
\hline
\multirow{2}*{instance} & \multirow{2}*{com.} & \multicolumn{2}{|c|}{MSGA} & \multicolumn{2}{|c}{ACO} & \multicolumn{2}{|c|}{Local search}\\
\cline{3-8}
 & & \multicolumn{1}{|c|}{$\overline{q}$} & \multicolumn{1}{|c|}{$\overline{t}$} & \multicolumn{1}{|c|}{$\overline{q}$} & \multicolumn{1}{|c|}{$\overline{t}$} & \multicolumn{1}{|c|}{$\overline{q}$} & \multicolumn{1}{|c|}{$\overline{t}$}\\
\hline
\multirow{3}*{mesh25x25.bb} & 62 & 36.95 & 546.854 & 31.1 & 880.551 & \textbf{38.85} & 1165.47\\
 & 156 & 44.65 & 863.007 & 47.5 & 965.921 & \textbf{55.5} & 1082.78\\
 & 250 & 50.5 & 672.962 & 60.5 & 972.396 & \textbf{67.95} & 967.087\\
\hline
\multirow{3}*{mesh15x15.bb} & 22 & 20.55 & 517.601 & 18.6 & 500.812 & \textbf{21} & 384.828\\
 & 56 & 27.15 & 651.27 & 28.35 & 988.782 & \textbf{30.3} & 485.693\\
 & 90 & 31 & 797.534 & 34.55 & 746.96 & \textbf{36.05} & 435.308\\
\hline
\multirow{3}*{bl-wr2-wht2.10-50.rand.bb} & 50 & 18.7 & 688.651 & 19.6 & 201.235 & \textbf{20.05} & 228.382\\
 & 125 & 27.2 & 643.51 & 31.15 & 338.446 & \textbf{31.2} & 241.047\\
 & 200 & 36.6 & 625.138 & 41.55 & 164.783 & \textbf{41.7} & 202.186\\
\hline
\multirow{3}*{bl-wr2-wht2.10-50.sdeg.bb} & 50 & 18.65 & 470.26 & 19.75 & 223.396 & \textbf{20.1} & 311.887\\
 & 125 & 28.1 & 662.916 & 31.55 & 163.151 & \textbf{31.85} & 357.25\\
 & 200 & 33.3 & 487.999 & 38.05 & 217.362 & \textbf{38.25} & 178.417\\
\hline
\end{tabular}

\end{center}
}
\end{small}
\caption{Experimental results of EDP problem}
\end{table*}

\subsubsection*{Acknowledgments}
We would like to thank Maria Jos\'{e} Blesa Aguilera who has kindly provided some graphs for the experimentation. This research is also partially supported by the Interuniversity Attraction Poles Programme (Belgian State, Belgian Science Policy).

\section{Bibliography}

\bibliographystyle{eptcs} 

\end{document}